\pdfoutput=1

\documentclass[11pt]{article}

\usepackage[final]{acl}

\usepackage{times}
\usepackage{latexsym}

\usepackage[T1]{fontenc}

\usepackage[utf8]{inputenc}

\usepackage{microtype}

\usepackage{inconsolata}

\usepackage{graphicx}

\usepackage{amsmath,amssymb} 
\usepackage{color}
\usepackage{algorithm}
\usepackage{algorithmic}
\usepackage{graphicx}
\usepackage{multirow}
\usepackage{subfigure}
\usepackage{comment}
\usepackage{booktabs}
\usepackage{stackengine}
\usepackage{xcolor,colortbl}
\usepackage{amsthm}     
\usepackage{caption}
\usepackage{wrapfig}
\usepackage{arydshln}
\usepackage[most]{tcolorbox}
\usepackage{lipsum}
\usepackage{algorithm}
\usepackage{algorithmic}
\usepackage{listings}

\usepackage{multirow} 
\usepackage{graphicx} 
\usepackage{booktabs} 

\definecolor{deepgreen}{rgb}{0,0.7,0}

\newcommand{\up}[1]{{\color{deepgreen} $\uparrow$\small{#1}}}
\newcommand{\down}[1]{\color{red} $\downarrow$\small{{#1}}}

\definecolor{codegray}{rgb}{0.5,0.5,0.5}
\definecolor{codepurple}{rgb}{0.58,0,0.82}
\definecolor{backcolour}{rgb}{0.95,0.95,0.92}
\lstdefinestyle{mystyle}{
    backgroundcolor=\color{backcolour},   
    basicstyle=\ttfamily\footnotesize,
    stringstyle=\color{codepurple},
    breakatwhitespace=false,         
    breaklines=true,                 
    captionpos=b,                    
    keepspaces=true,             
    showspaces=false,                
    showstringspaces=false,
    showtabs=false,                  
    tabsize=2,
    breakautoindent=true, 
    breakindent=0pt
}
\title{Logic-of-Thought: Injecting Logic into Contexts for Full Reasoning in Large Language Models}

\author{
Tongxuan Liu$^{1,4}$\footnotemark[1]\footnotemark[2],\ Wenjiang Xu$^3$\footnotemark[1],\ Weizhe Huang$^{1,4}$,\ Yuting Zeng$^1$,\ Jiaxing Wang$^4$, \\
\bf Xingyu Wang$^3$,\ Hailong Yang$^2$,\ Jing Li$^1$\footnotemark[2]\\
$^1$ University of Science and Technology of China, \quad $^2$ Beihang University, \\
$^3$ Institute of Automation, Chinese Academy of Sciences, \quad $^4$ JD.com \\
\texttt{\{tongxuan.ltx, hwz871982879, yuting\_zeng\}@mail.ustc.edu.cn},\\
\texttt{\{xuwenjiang2024, wangxingyu2024\}@ia.ac.cn},\\
\texttt{lj@ustc.edu.cn, hailong.yang@buaa.edu.cn, wangjiaxing41@jd.com} \\
}

\begin{document}

\maketitle
\footnotetext[1]{Equal contributions.}
\footnotetext[2]{Corresponding authors.}
\begin{abstract}
Large Language Models (LLMs) have demonstrated remarkable capabilities across various tasks but their performance in complex logical reasoning tasks remains unsatisfactory. Although some prompting methods, such as Chain-of-Thought, can improve the reasoning ability of LLMs to some extent, they suffer from an unfaithful issue where derived conclusions may not align with the generated reasoning chain. To address this issue, some studies employ the approach of propositional logic to further enhance logical reasoning abilities of LLMs. However, the potential omissions in the extraction of logical expressions in these methods can cause information loss in the logical reasoning process, thereby generating incorrect results. To this end, we propose Logic-of-Thought (LoT) prompting which employs propositional logic to generate expanded logical information descriptions and utilizes them as an additional augmentation to original contexts, thereby ensuring information completeness and enhancing logical reasoning ability. LoT is orthogonal to existing prompting methods and can be seamlessly integrated with them. Extensive experiments demonstrate that LoT boosts the performance of various prompting methods with a striking margin across five logical reasoning tasks. In particular, LoT enhances Chain-of-Thought's performance on the ReClor dataset by +4.35\%, improves Chain-of-Thought with Self-Consistency's performance on the RuleTaker dataset by +3.52\%, and boosts performance of Tree-of-Thoughts on the ProofWriter dataset by +8\%\footnote[3]{Code and data are available at \url{https://github.com/HEA1OR/lot}.}. 

\end{abstract}

\section{Introduction}
\label{sec:introduction}

In recent years, Large Language Models (LLMs) have demonstrated excellent capabilities across various NLP tasks \cite{openai2024gpt4,anil2023palm,touvron2023llama2}. However, even the most advanced LLMs exhibit limited performance in mathematics and complex logical reasoning tasks \cite{arkoudas2023gpt,liu2023evaluating}. Chain-of-Thought (CoT) prompting \cite{kojima2022large,wei2023chainofthought,nye2021show} has emerged as a promising approach to improve logical reasoning capabilities, which enhances reasoning abilities by adding intermediate steps in the reasoning process. 
Subsequent research has sought to simulate human reasoning processes by expanding the Chain-of-Thought into more complex reasoning topology. For example, Tree-of-Thoughts (ToT) \cite{yao2024tree} extends into a tree-like reasoning topology, exploring more reasoning branches at each step and supporting backtracking. 
STaR \cite{eric2022star} and Chain-of-Thought with Self-Consistency (CoT-SC) \cite{wang2022self} generate multiple chains of thought or reasoning paths, selecting the most optimized and consistent answers from these. However, \cite{bao2024llms,lanham2023measuring,lyu2023faithful,turpin2024language} observe that LLMs occasionally exhibit unfaithful reasoning, wherein the derived conclusions do not adhere to the previously generated reasoning chain.

\begin{figure*}[t]
    \centering
    \includegraphics[width=1\textwidth]{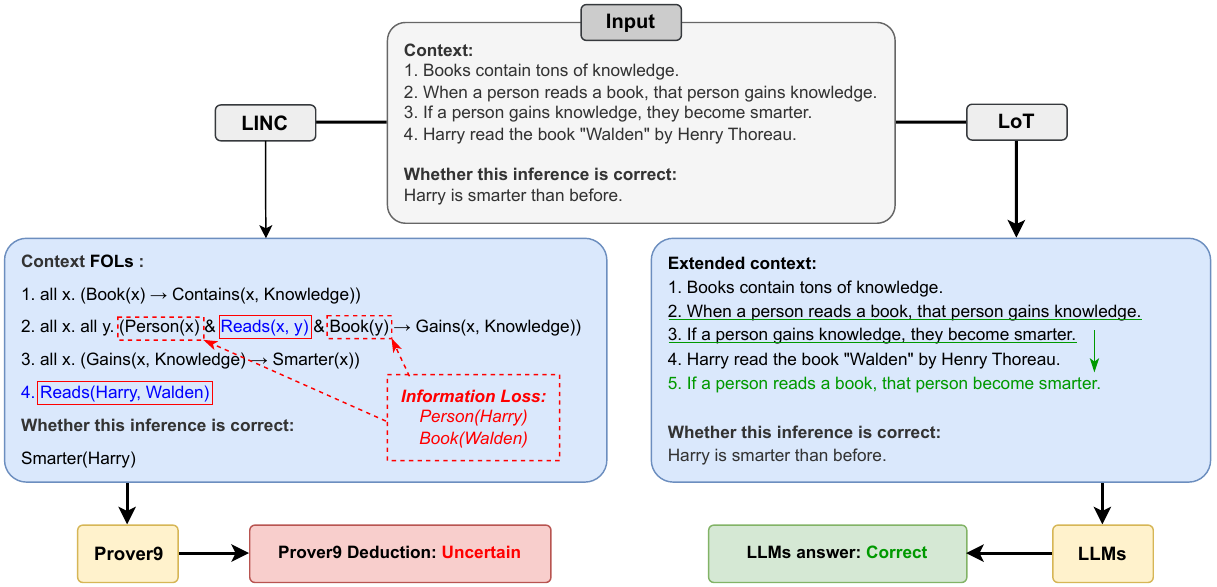}
    \caption{\textbf{Comparison between LINC and LoT.} The left part involves the workflow of LINC, which converts problems into logical expressions and then derives results using Prover9. However, LINC loses fact information \textit{Person(Harry)} and \textit{Book(Walden)}, leading to errors. On the right side, LoT generates supplementary logical information \textit{"If a person reads a book, that person become smarter"} that is seamlessly integrated into the original prompt, thereby enhancing the capability of LLMs to produce accurate results.}
    \label{fig1}
\end{figure*}

To tackle the challenge of the unfaithfulness in the reasoning process, researchers have proposed many neuro-symbolic methods that integrate LLMs with symbolic reasoning, such as Faithful Chain-of-Thought \cite{lyu2023faithful}, LINC \cite{olausson2023linc}, Logic-LM \cite{pan2023logic} and SatLM \cite{ye2024satlm}. These methods follow a similar process: Initially, the problem and objectives are translated into symbolic expressions. Subsequently, symbolic results are derived through external tools such as symbolic solvers. Finally, it's optional to explain symbolic results using LLMs or interpreters. However, these existing neuro-symbolic methods inevitably suffer from the issue of information loss, which results from omissions in the extraction of logical expressions and directly leads to incorrect intermediate reasoning processes. As illustrated in the Figure \ref{fig1}, in the extraction process of logical expressions in LINC, two key pieces of hidden information \textit{"Harry is a person"} and \textit{"Walden is a book"} are lost, which makes it impossible for the symbolic solver Prover9 to obtain the correct reasoning result.

To address the issue of information loss, in this paper, we propose a novel zero-shot prompting method named Logic-of-Thought (LoT). Specifically, LoT first extracts propositions and logical expressions from the input context, expands these logical expressions according to logical reasoning laws, and converts the deduced logical expressions back into natural language form. Then LoT considers these extended logical descriptions as additional logical augmentation for LLMs and concatenates it with the original context, which not only encourages LLMs to utilize these new deduced logical information when answering the original question but also ensures information completeness through preserving full original contexts for LLMs reasoning, thereby enhancing logical reasoning ability.
Additionally, the LoT prompting approach is compatible and orthogonal to existing prompting methods, enabling seamless integration of these methods. To validate the effectiveness of LoT, we conduct extensive experiments to evaluate its capability in boosting various prompting methods such as CoT, SC, CoT-SC and ToT across five logical reasoning datasets. Experimental results demonstrate that LoT prompting can seamlessly integrate with existing prompting methods and significantly boost their performance in logical reasoning. Specifically, LoT significantly enhances the performance of CoT on the ReClor dataset, achieving an improvement in accuracy up to +4.35\%. Furthermore, LoT improves the SC's performance on the ReClor dataset by a remarkable +6.52\%. Moreover, LoT boost the accuracy of CoT-SC on RuleTaker by +3.52\%. Additionally, LoT effectively elevates the performance of ToT on the ProofWriter dataset, resulting in a significant improvement of +8\%. 

The main contributions of this paper are as follows:

\begin{itemize}
    \item[1.] We propose a novel prompting method Logic-of-Thought (LoT) to address the issue of information loss in existing neuro-symbolic methods by generating logical proposition descriptions as augmentations for original prompts.
    \item[2.] We integrate LoT with a variety of distinct prompting techniques, including Chain-of-Thought (CoT), Self-Consistency (SC), Chain-of-Thought with Self-Consistency (CoT-SC), Tree-of-Thoughts (ToT), by leveraging the orthogonal capabilities of LoT. 
    \item[3.] We conduct extensive experiments to evaluate the effectiveness of LoT in enhancing the capabilities of different prompting techniques across diverse logical reasoning tasks. The results demonstrate the significant effectiveness of LoT in boosting the performance of various prompting methods.

\end{itemize}

\section{Preliminary}
\label{sec:preliminary.tex}

As this study focuses on logical reasoning tasks, we first provide some definitions and symbols about the propositional logic system, which will be used throughout the paper.
\begin{itemize}
\item \textit{Propositions} are defined as declarative sentences that have clear truth-value characteristic and cannot be simultaneously true and false. In this context, propositions are considered fundamental elements of logical expressions. We use standard uppercase letters such as $A$, $B$, $C$ to symbolize specific propositions, exemplified by statements like \textit{"you have keyboarding skills"}, and lowercase letters such as $p$, $q$, $r$ to refer to any proposition. 

\item \textit{Connectives} are defined as operators on propositions, which can operate on a single proposition or link propositions together to form a new \textit{logical expression}. In this study, We mainly focus on three connectives: $\neg$, $\to$ and $\wedge$. Herein, negative $\neg$ denotes the negation operation for a specific logical symbol (e.g., $\neg p$). Implication $\to$ signifies a sufficient condition or causal relationship between two propositions (e.g., $p \to q$). Conjunction $\wedge$ also operates on two propositions, which represents that the entire expression is true only if both propositions are true (e.g., $p \wedge q$). 
\item \textit{Logical reasoning laws} are defined as the deducing relation between two logical expressions. 
In this study, we utilize three basic logical reasoning laws: the Double Negation Law $\neg \neg p \Leftrightarrow p $, the Contraposition Law $(p \to q) \Leftrightarrow (\neg q \to \neg p) $, and the Transitive Law $(p \to q) \wedge (q \to r) \Rightarrow (p \to r) $, which all align with human intuition and are fundamental and widely used in propositional logic \cite{buning1999propositional}.
\end{itemize}
Although the presented logic system setting is straightforward, our paper primarily concentrates on introducing a new prompting paradigm to address information loss in existing neuro-symbolic methods. Moreover, notable enhancements have already been achieved within this setting (See Section \ref{main_results}). Therefore, we leave the exploration of incorporating more diverse connectives and laws in our method to future work.
\section{Methodology}
\label{sec:methodology}

\begin{figure*}[t]
    \centering
    \includegraphics[width=1\textwidth]{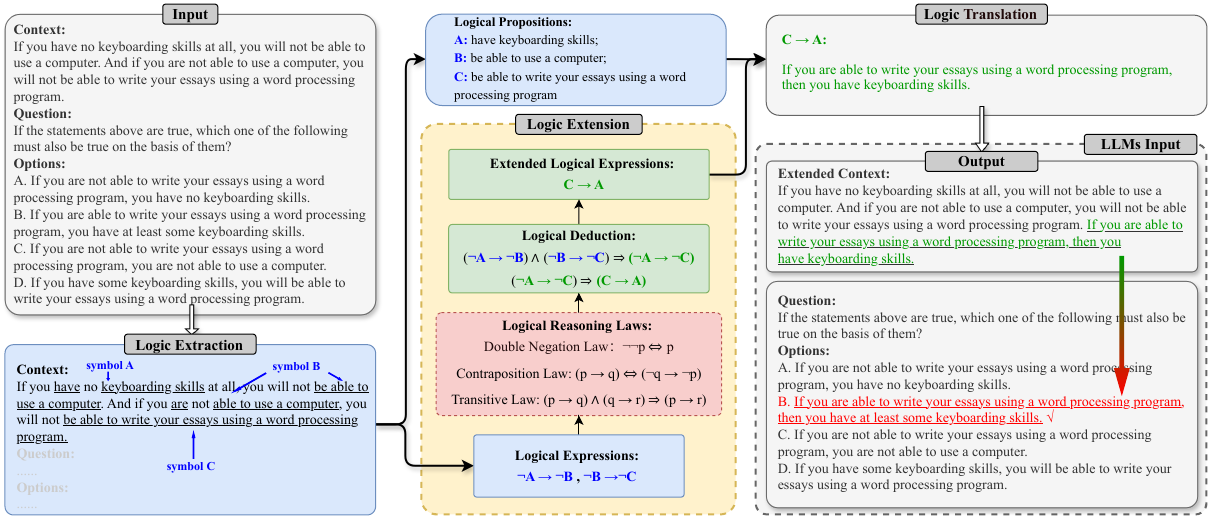}
    \caption{\textbf{The framework of LoT consisting of three phases.} On the left side of the diagram is the Logic Extraction phase, where we employ LLMs to extract propositions and logical relations. In the middle is the Logic Extension phase, where we apply logical reasoning laws to derive logical expressions. On the right side is the Logic Translation phase, where we utilize LLMs to translate logical expressions into their natural language descriptions.} 
    \label{fig2}
\end{figure*}

\paragraph{Overview.} Figure \ref{fig2} presents an overview of LoT, which consists of three phases. Firstly, in the Logic Extraction phase, propositions and logical relations are extracted from the input context using LLMs to output logical expressions. Secondly, in the Logic Extension phase, the logical expressions are expanded through Python-implemented logical rules. Thirdly, in the Logic Translation phase, the expanded logical expressions are translated into natural language descriptions of logical information through LLMs. And then, the logical information is incorporated into the input prompt, forming a comprehensive and novel input prompt for LLMs. The following sections provide detailed introduction to the phases of Logic Extraction, Logic Extension, and Logic Translation. 

\paragraph{Logic Extraction.} In the Logic Extraction phase, we use LLMs to extract formal logic expressions from the input context through two stages. Firstly, we instruct LLMs to select sentences containing conditional reasoning relationships from the input context to generate collection of sentences with logical relationships. Subsequently, we use LLMs to extract the set of propositional symbols $\mathcal{P}$ and the set of logical expressions $\mathcal{E}$ from the collection. During the process of Logic Extraction, LLMs identify propositions with similar meanings and represent them using identical propositional symbols. Furthermore, LLMs analyze the logical relationships between propositions from their natural language descriptions, ultimately deriving the logical expressions. For propositions expressing opposite meanings, the negation $\neg$ is added. When there is a conditional relationship between two propositions, the implication $\to$ is used to connect their corresponding propositional symbols. We also incorporate well-designed hints about logical relationships into the prompt, such as phrases like "if...then..." or "...causes..." to further guide LLMs in analyzing logical connections and minimize errors. For example, as depicted in Figure \ref{fig2}, LLMs extract the same meaning description \textit{"be able to use a computer"} from two different sentences, symbolized as $B$. Then, through analyzing its logical relationship with other propositions, LLMs apply $\neg$ to $B$ and another proposition $A$ and add $\to$ between them, which results in a new logical expression $\neg A \to \neg B$.

\paragraph{Logic Extension.} During the Logic Extension phase, we apply logical reasoning laws to the collection of logical expressions from the Logic Extraction phase. These logical expressions can be further expanded using a Python program to implement logical deduction. As illustrated in the Figure \ref{fig2}, the extracted logical expressions $\neg A \to \neg B$ and $\neg B \to \neg C$ serve as inputs for our logical deduction program. Through expansion based on Transitive Law and Contraposition Law, we finally obtain the new expression $C\to A$, which will be used in the next phase.

\paragraph{Logic Translation.} During the Logic Translation phase, we use LLMs to translate the generated extended logical expressions into natural language descriptions. Subsequently, we combine the natural language descriptions of propositional symbols according to the extended logical expressions to form a new part of the original input prompt. Through this approach, we inject the deduced logical information as additional augmentation into the original prompt, thus avoiding information loss. As shown in Figure \ref{fig2}, by associating $C$ with its description \textit{"be able to write your essays using a word processing program"}, $A$ with its description \textit{"have keyboarding skills"}, and $\to$ with the logical description \textit{"if...then..."}, we can translate the aforementioned logical expression $C\to A$ back to its natural language description and add it to original prompts as new input prompts.

\section{Experiments}
\label{sec:experiments}
\begin{table*}[t] 
\renewcommand{\arraystretch}{1.5}
\centering
\resizebox{\textwidth}{!}{
\begin{tabular}
{c|lllll|lllll}
\toprule
    \multirow{2}{*}{\textbf{Method}} & \multicolumn{5}{c|}{\textbf{\texttt{GPT-3.5-turbo-0125}}} & \multicolumn{5}{c}{\textbf{\texttt{GPT-4-0613}}} \\ \cline{2-11}
    & \textbf{ReClor} & \textbf{LogiQA} & \textbf{RuleTaker} & \textbf{ProofWriter} & \textbf{FOLIO} & \textbf{ReClor} & \textbf{LogiQA} & \textbf{RuleTaker} & \textbf{ProofWriter} & \textbf{FOLIO} \\ \midrule
Direct & 46.20 & 36.44 & 51.89 & 52.87 & 68.89 & 72.17 & 59.22 & 64.30 & 63.74 & 82.96 \\
LoT & 56.02~\up{9.82} & 36.85~\up{0.41} & 59.44~\up{7.55} & 59.35~\up{6.48} & 76.00~\up{7.11} & 77.98~\up{5.81} & 60.11~\up{0.89} & 64.65~\up{0.35} & 65.58~\up{1.84} & 83.55~\up{0.59} \\
\midrule
CoT & 52.17 & 39.75 & 60.56 & 61.02 & 81.19 & 77.39 & 58.97 & 68.69 & 69.83 & 85.33 \\
LoT + CoT & 56.52~\up{4.35} & 41.20~\up{1.45} & 62.46~\up{1.90} & 63.35~\up{2.33} & 78.96~\down{2.23} & 79.13~\up{1.74} & 58.40~\down{0.57} & 69.02~\up{0.33} & 70.56~\up{0.73} & 85.48~\up{0.15} \\
\midrule
SC(5) & 56.52 & 37.10 & 52.43 & 53.91 & 70.37 & 73.91 & 59.98 & 64.32 & 64.16 & 82.96 \\
LoT + SC(5) & 58.70~\up{2.18} & 37.48~\up{0.38} & 59.98~\up{7.55} & 60.51~\up{6.60} & 77.04~\up{6.67} & 80.43~\up{6.52} & 60.75~\up{0.77} & 64.53~\up{0.21} & 65.99~\up{1.83} & 82.96~\up{0.00} \\
\midrule
CoT-SC(5) & 58.70 & 41.86 & 61.63 & 62.54 & 81.48 & 80.43 & \underline{\textbf{61.67}} & 69.49 & 70.56 & 86.67 \\
LoT + CoT-SC(5) & \underline{\textbf{60.87}}~\up{2.17} & \underline{\textbf{42.63}}~\up{0.77} & \underline{\textbf{65.15}}~\up{3.52} & \underline{\textbf{65.89}}~\up{3.35} & \underline{\textbf{81.48}}~\up{0.00} & \underline{\textbf{82.61}}~\up{2.18} & 60.29~\down{1.38} & \underline{\textbf{70.73}}~\up{1.24} & \underline{\textbf{71.98}}~\up{1.42} & \underline{\textbf{88.15}}~\up{1.48} \\

\bottomrule
\end{tabular}
}
\caption{\textbf{Main results of combining LoT with various prompting methods.} The number in green indicates an enhancement in performance, while the number in red signifies a decline in performance. 
}

\label{tbl:main_result1}
\end{table*}
\subsection{Datasets}

In the experiment, we employ five logical reasoning datasets: (1) ReClor \cite{yu2020reclor}, which is collected from standardized test logical reasoning questions, including the Law School Admission Test (LSAT) and the Graduate Management Admission Test (GMAT); (2) LogiQA \cite{liu2020logiqa}, which is derived from expert-written questions for testing human logical reasoning; (3) RuleTaker \cite{clark2021transformers}, which is automatically generated via programming, utilizing connectives including $\wedge$, $\neg$, and $\to$; (4) ProofWriter \cite{tafjord2021proofwriter}, which comprises numerous small rulebases composed of facts and rules; and (5) FOLIO \cite{han2022folio}, which is characterized by its human annotations and first-order logic annotations. 


\subsection{Baselines} 

We consider 5 widely used prompting methods and 2 neuro-symbolic methods for comparison. The prompting methods include: (1) Direct prompting, which directly input the question; (2) Self-Consistency (SC) \cite{wang2022self}, which employs majority voting to aggregate responses from multiple Direct prompting and defaults to 5 times; (3) CoT \cite{kojima2022large,wei2023chainofthought,nye2021show}, which utilizes a progressive thinking approach for reasoning; (4) CoT-SC \cite{wang2022self}, which applies majority voting to aggregate multiple CoT and also defaults to involving 5 reasoning paths; (5) ToT \cite{yao2024tree}, which models the reasoning process as a thought search tree. We also choose two recent neuro-symbolic methods, SatLM \cite{ye2024satlm} and LINC \cite{olausson2023linc}, which leverage automated theorem provers to assist LLMs in reasoning. 

\subsection{Experiment Setup}


\paragraph{Main experiments.} Main experiments employ four prompting methods including Direct, CoT, SC, CoT-SC and combination of these prompting methods with LoT using \texttt{GPT-3.5-turbo-0125} and \texttt{GPT-4-0613} across five datasets. We utilize the zero-shot setting for all methods and employ default values for temperature \textit{top\_p} and \textit{top\_k}. For the ReClor dataset, we selected all 46 data entries in the test set that pertain to the implication section. For the ProofWriter dataset, we selected all 985 test data points that conform to the Closed-World Assumption (CWA) and have a depth of 5. For the RuleTaker dataset, we also selected all 967 test data points that conform to CWA and have a depth of 5. For the LogiQA dataset, we selected a combined set of 1302 Chinese and English test data points. For the FOLIO dataset, we selected 135 test data entries that conform to CWA.

\paragraph{Comparison between LoT and existing neuro-symbolic methods.} We conduct comparative experiment with SatLM and LINC. To leverage the SatLM implementation \cite{ye2024satlm}, we reran it in a one-shot setting with \texttt{GPT-3.5-turbo-instruct}. To ensure compatibility with our experimental setup, we also chose to conduct evaluation on the ReClor dataset, as selected 46 data entries in the ReClor dataset closely mirror the LSAT dataset tested in \cite{ye2024satlm}. For LINC, we re-evaluated its performance on the FOLIO dataset through using its public code and retaining all its original settings (e.g., 8-shot).

\paragraph{Analysis of ToT with LoT.} In this experiment, we evaluate the performance enhancement achieved by LoT under the guidance of ToT on the ProofWriter dataset, leveraging \texttt{GPT-4-0613}. We re-implemented ToT based on the work presented in \cite{zhang2024cumulative}. In this experiment, the Direct, ToT, and LoT+ToT were all implemented in few-shot setting. For the ToT-related experiments, each successful state explored up to five new states. The success or failure of a state was verified by assessing its compliance with the established rules. The exploration process terminated either after achieving four successful state explorations or when no new states were available for exploration. We utilized 100 randomly selected data entries with a depth of 5 from the ProofWriter validation set.


\subsection{Main Results}
\label{main_results}

In this section, we integrate LoT prompting with four baseline prompting methods, namely Direct, CoT, SC and CoT-SC, to conduct a comparative analysis of whether LoT enhances logical reasoning abilities across five distinct datasets. The results presented in Table \ref{tbl:main_result1} reveal some key observations:

\begin{itemize}

\item Combining LoT with existing prompting methods can achieve best performance, which highlights the superiority of our methods. Specifically, LoT+CoT-SC(5) outperforms all other methods across all five datasets with \texttt{GPT-3.5-turbo-0125} and four datasets with \texttt{GPT-4-0613}. 

\begin{figure}[t]
    \centering
    \includegraphics[width=0.45\textwidth]{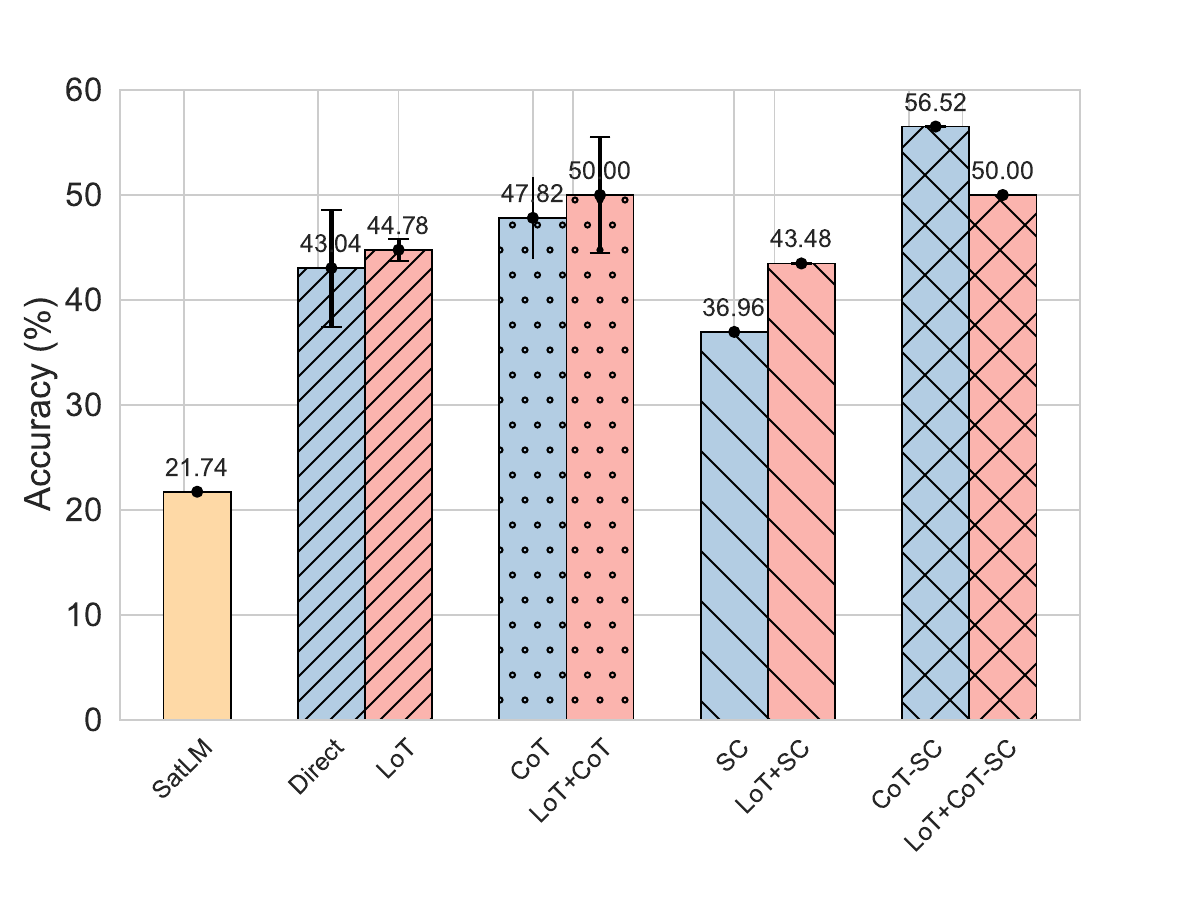}
    \caption{\textbf{Comparison between SatLM and LoT in the ReClor dataset.} }
    \label{fig4}
\end{figure}

\item LoT enhances the performance of four baseline prompting methods in most experiments, suggesting that LoT can be seamlessly integrated into existing prompting methods to further improve the logical reasoning ability of LLMs.
Among total 40 comparisons (including four baseline prompting methods across five datasets with two LLMs), LoT significantly enhances the performance of baseline prompting methods in 37 instances. 

\begin{figure*}[t]
    \centering
    \includegraphics[width=1\textwidth]{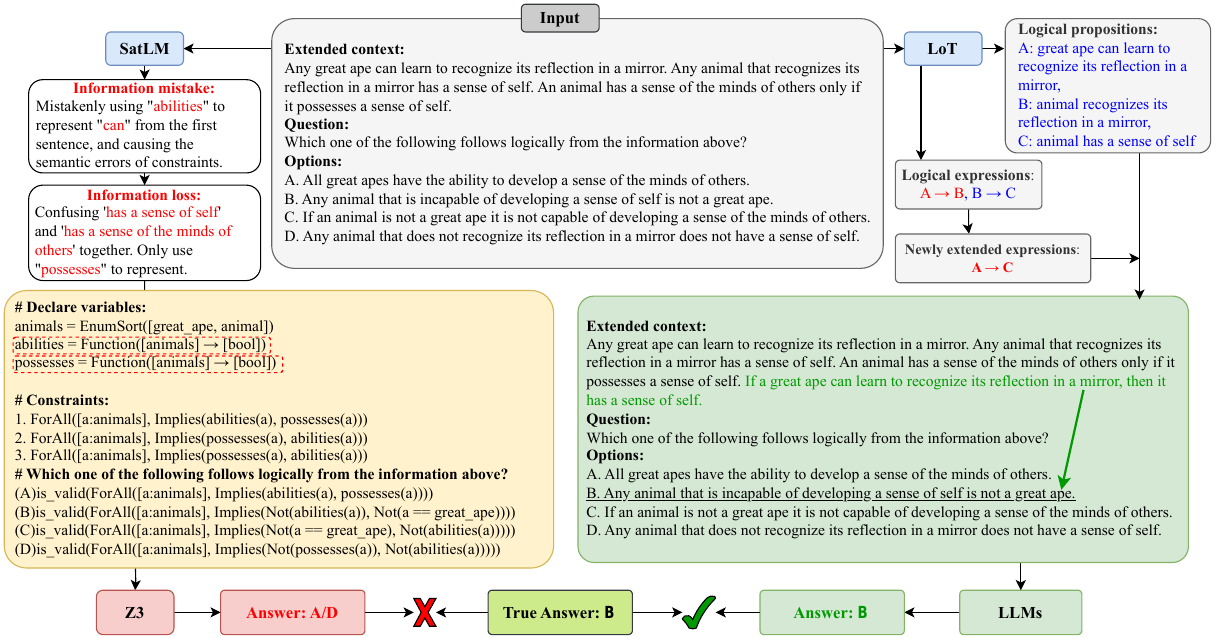}
    \caption{\textbf{A comparative case of SatLM and LoT from the ReClor dataset.} }
    \label{fig:satlm_lot}
\end{figure*}

\item Upon utilizing GPT-4 on the LogiQA dataset, we observe that LoT+CoT and LoT+CoT-SC marginally trailed behind CoT and CoT-SC, recording a decline of 0.57\% and 1.38\% respectively. We find that the primary factor is the deviation in extracting logical information during the Logic Extraction phase, where we provide an example in the Appendix \ref{appendix:limitation-lot}. 

\item LoT achieves significant enhancements in the accuracy of Direct across all datasets and outperforms CoT in eight out of ten sets of comparative data. Thus, this provides compelling evidence that the standalone utilization of LoT can achieve or even exceed the logical reasoning capability exhibited by CoT.

\end{itemize}

\subsection{Comparison with Neuro-symbolic Methods}

In this section, we conduct comparison with two neuro-symbolic approaches, SatLM and LINC. 
\paragraph{Performance Study.} As shown in Figure \ref{fig4}, it can be first observed that LoT significantly outperforms SatLM in terms of accuracy on the Reclor dataset as well as obtains notable improvements across various prompting methods, including Direct (+1.74\%), CoT (+2.18\%), and SC (+6.52\%), which also shows LoT's effectiveness. But we observe that in this set of experiments, the performance of LoT+CoT-SC is inferior to that of CoT-SC. We speculate that this is due to a bias in the extraction of logical information, which is also discussed in Section \ref{main_results}. 
Additionally, from Table \ref{tbl:comp_linc}, we can also observe that LoT can significantly outperform LINC, which also shows LoT's effectiveness.
Furthermore, SatLM and LINC exhibit poor performance under our setting compared to Direct prompting. This is in line with our motivation that these neuro-symbolic methods are more likely to encounter the issue of information loss when extracting logical symbolic expressions, po compromising their overall performance. 

\begin{table}[t] 
\renewcommand{\arraystretch}{1}
\centering
\resizebox{\linewidth}{!}{
\begin{tabular}{ccc}
\toprule
\textbf{Method} & \textbf{\texttt{GPT-3.5-turbo-0125}} & \textbf{\texttt{GPT-4-0613}} \\ 
\midrule
Direct & 68.89 & 82.96 \\
LINC & 45.19 & 55.56 \\ 
LoT & \underline{\textbf{76.00}} & \underline{\textbf{83.55}} \\

\bottomrule
\end{tabular}
}
\caption{\textbf{Comparison between LINC and LoT in the FOLIO dataset.}}
\label{tbl:comp_linc}
\end{table}

\paragraph{Case Study.} We also conduct a comparative case study between SatLM and LoT. As depicted in Figure \ref{fig:satlm_lot}, SatLM induces information mistakes and loss. Specifically, during logical extraction, SatLM erroneously employs \textit{"abilities"} to represent \textit{"can"}, leading to semantic errors in constraints. Additionally, SatLM confuses \textit{"has a sense of self"} with \textit{"has a sense of the minds of others"} and only utilizes \textit{"possesses"} to represent them together. In contrast to SatLM, LoT successfully extracts logical proposition descriptions and symbolizes them. Here, we have an interesting finding: when directly examining the extracted logical expressions, a small mistake in $A \rightarrow B$ results in an incorrect $A \rightarrow C$ (i.e., we cannot infer general \textit{"animal"} from specific \textit{"great ape"}). However, when translating the deduced logical expressions $A \rightarrow C$ into natural language, LLMs recognize the subordinate relationship between \textit{"ape"} and \textit{"animal"} and correct this error, resulting in correct augmentation to prompts and right answers. This reflects that LoT fully leverages the LLM's understanding of natural language descriptions, enabling it to correct errors from earlier phases in the three-phase process. This avoids the pitfalls of neuro-symbolic methods, which rely entirely on the accuracy of logical extraction, where errors in intermediate results directly propagate to errors in the final outcome.

\begin{figure}[t]
    \centering
    \includegraphics[width=0.45\textwidth]{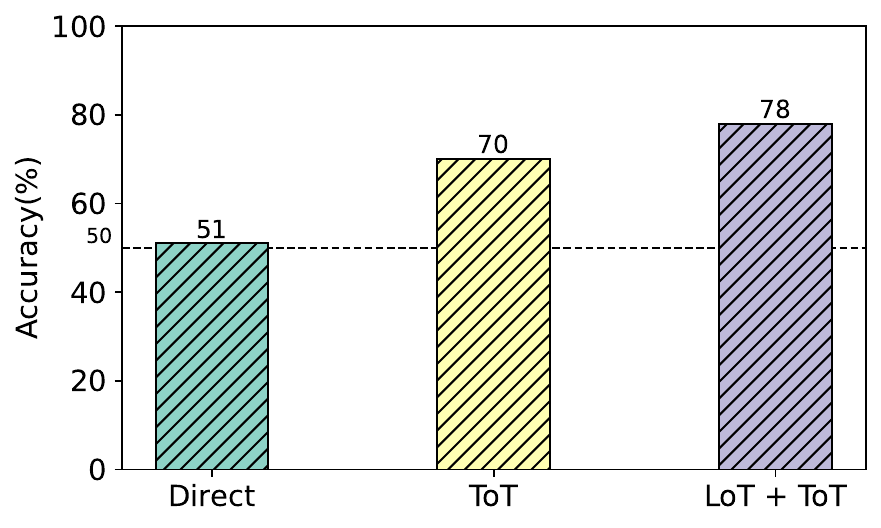}
    \caption{\textbf{Comparison between ToT and LoT+ToT in the ProofWriter dataset.} We conducted this experiments based on \cite{zhang2024cumulative}.}
    \label{fig5}
\end{figure}



\subsection{In-depth Analysis of ToT with LoT}

In this experiment, we assess the enhancing effect of LoT on ToT. As shown in Figure \ref{fig5}, we can observe that under the complex reasoning scenario with a deduction depth of 5 in the ProofWriter dataset, Direct only achieves performance similar to random guessing (50\%). The accuracy rate of ToT is +19\% higher than Direct, reaching 70\%, which shows that ToT can assist LLMs in better solving multi-step reasoning. LoT+ToT reaches +8\% increase in accuracy compared to ToT, indicating that LoT can effectively enhance the ability of ToT in complex logical reasoning.

To further investigate the influence of LoT on ToT, we carefully analyze a range of indices within ToT, including the count of total states (TS), the count of successful states (SS), and the proportion of samples that complete full reasoning (FS), which means four successful states explorations are achieved. As shown in Table \ref{tbl:state}, we observe +2.14\% increase in the overall states of LoT+ToT compared to ToT. This suggests that LoT facilitates an expanded exploration scope for ToT. Moreover, LoT improves the full reasoning of ToT by +2\%, which shows that LoT has a more comprehensive explored space. Furthermore, compared to ToT, LoT+ToT also exhibits a +5.06\% increase in successful states, indicating that LoT can significantly enhance the effectiveness of ToT's explored states. 
We present an example in Appendix \ref{appendix:tot-vs-lot+tot}, comparing state exploration between ToT and LoT+ToT.

\begin{table}[t] 
\centering
\resizebox{\linewidth}{!}{
\begin{tabular}
{clll}
\toprule
\textbf{Method} & \textbf{TS} & \textbf{SS} &\textbf{FR (\%)} \\ 
\midrule
ToT & 18.70 & 7.70 & 90 \\
LoT+ToT & \underline{\textbf{19.10}}~\up{2.14\%} & \underline{\textbf{8.09}}~\up{5.06\%} & \underline{\textbf{92}}~\up{2\%} \\

\bottomrule
\end{tabular}
}
\caption{\textbf{Comparison of reasoning states.} TS means the count of total states. SS means the count of successful states. FR means the proportion of samples that complete full reasoning.}
\label{tbl:state}
\end{table}


\subsection{Ablation Study}
In this section, we investigate the impact of the quality of generated additional logical information on the results. Specifically, we remove one of the three logical reasoning laws (i.e., the Contraposition Law in our experiment) and retain only half of the generated information. These two conditions are, respectively, denoted as \textit{w/o Contraposition Law} and \textit{w/o Half of Generated Description}. We conduct the ablation experiment using \texttt{GPT-4-0613} in the FOLIO dataset.
As shown in Table \ref{tbl:ablation}, either of the variants reduces additional logical information generated by LoT, ultimately leading to a decrease in accuracy. This underscores the effectiveness of the logical information deduced by LoT.

\begin{table}[t] 
\centering
\resizebox{\linewidth}{!}{
\begin{tabular}
{lc}
\toprule
\textbf{Method} & \textbf{FOLIO} \\ 
\midrule
LoT &  \underline{\textbf{76.00}} \\
\midrule
w/o Contraposition Law &  72.22 \\
w/o Half of Generated Description &  72.96 \\

\bottomrule
\end{tabular}
}
\caption{\textbf{Ablation study of generated additional logical information.} \textit{w/o Contraposition Law} refers to removing Contraposition Law during the Logic Extension phase. \textit{w/o Half of Generated Description} means retaining only half of the additionally generated logical descriptions.}
\label{tbl:ablation}
\end{table}

In addition, we also conduct another ablation experiment of the Logic Extension phase on the LogiQA and RuleTaker datasets. Specifically, we directly translate the logical expressions extracted from the Logic Extraction phase into natural language without performing logical extension (i.e. removing the Logic Extension). As shown in Table \ref{tbl:ablationofextension}, the results demonstrate that removing the Logic Extension leads to a drop in LoT's accuracy, though it still outperforms the Direct approach. This highlights the necessity of first extracting and analyzing logical relationships before allowing the LLM to directly address the problem, as well as the effectiveness of leveraging logical reasoning laws to extend the extracted logical expressions and enhance the injected logical information.

\begin{table}[t] 
\centering
\resizebox{\linewidth}{!}{
\begin{tabular}
{lcc}
\toprule
\textbf{Method} & \textbf{LogiQA} & \textbf{RuleTaker}\\ 
\midrule
Direct &  36.44 & 51.89 \\

LoT &  \underline{\textbf{36.85}} & \underline{\textbf{59.44}}\\

LoT w/o Extension &  36.79 & 59.05 \\
\bottomrule
\end{tabular}
}
\caption{\textbf{Ablation study of Logic Extension.}}
\label{tbl:ablationofextension}
\end{table}

\section{Related Work}
\label{sec:related-work.tex}

\subsection{Prompting Methods for LLMs Reasoning}

Numerous studies are dedicated to exploring enhancements in LLMs reasoning through prompting methods \cite{kojima2022large, wei2023chainofthought, wang2022self}. CoT \cite{wei2023chainofthought}, which breaks down a multi-step reasoning problem into multiple intermediate steps to gradually generate answers, has significantly improved logical reasoning, mathematical logic, and interpretability. 
CoT-SC \cite{wang2022self} further generates multiple thought chains, and the final answer is obtained through majority voting. Least-To-Most \cite{zhou2022least} deconstructs a problem into multiple sub-questions, addressing them step by step, with the answer to the previous sub-question serving as the input for the next. Similar decomposition methods of sub-problems include Lambada \cite{kazemi2022lambada} and the Divide-and-Conquer \cite{zhang2024guiding}. \cite{lightman2023lets} employs a process-supervised method, providing feedback on the intermediate reasoning process to enhance logical reasoning capabilities. \cite{shum2023automatic,eric2022star,zhou2022large} utilize various strategies to select optimal candidates from multiple chains of thought. ToT \cite{yao2024tree} and GoT \cite{besta2024graph} achieve logical branching and the aggregation of multiple thoughts by utilizing more complex reasoning topology.
However, these prompting methods occasionally exhibit unfaithful reasoning and lack in-depth exploration of logical information in logical reasoning tasks.
\subsection{Neuro-symbolic Methods}

The neuro-symbolic methods, which combine LLMs with symbolic reasoning, are considered an effective approach to address the issue of unfaithful reasoning and enhance the logical reasoning ability of LLMs \cite{wan2024b, olausson2023linc,ye2024satlm}. LReasoner \cite{wang2022logic} proposes a framework for context extension that expands the logical information contained in the context by applying logical reasoning laws. 
Logic-LM \cite{pan2023logic} initially utilizes LLMs to transform natural language problems into symbolic formulas. Subsequently, a symbolic solver is determined to reason about the formalized problems. Moreover, a self-refinement module is introduced, which utilizes error messages from the symbolic solver to modify the symbolic formalization. SatLM \cite{ye2024satlm} utilizes LLMs to generate declarative task specifications rather than imperative programs, and leverages readily available automated theorem solver to derive the final answers. LINC \cite{olausson2023linc} considers LLMs as a semantic parser, which translates premises and conclusions from natural language into first-order logic expressions, which are then offloaded to an external theorem solver.
However, these neuro-symbolic methods rely entirely on symbolic solvers, which inherently leads to information loss in extracting logical expressions and limits their accuracy.
\section{Conclusion}
\label{sec:conclusion}

In this paper, we introduce a zero-shot prompting approach Logic-of-Thought (LoT), designed to address the challenge of information loss inherent in existing neuro-symbolic methods. LoT leverages propositional logic to derive expanded logical information from input context, which serves as a supplementary augmentation to the original prompts, and can enhance logical reasoning capabilities of LLMs. LoT exhibits compatibility with widely used prompting techniques. In the experiments, we demonstrate that LoT significantly boosts the performance of various existing prompting methods across multiple logical reasoning datasets and can be seamlessly integrated with them. 
\section{Limitations}
\label{sec:limitations.tex}

Although our proposed LoT has achieved excellent performance in various logical reasoning tasks, there are still some limitations in LoT. Firstly, current LoT supports a limited set of connectives and logical reasoning laws. More connectives and logical reasoning laws in LoT means more complex prompt design in the Logic Extraction and Logic Translation phase, and increased difficulty in logical deducing in the Logic Extension phase. In the future, we will try to include additional connectives and logical reasoning laws in LoT to further enhance the logical reasoning capabilities. 

Additionally, although LoT preserve original question structures and utilizes extra deduced logical information as additional augmentation to mitigate information loss issue, hallucination issues inherent in LLMs can still lead to some failure in the Logic Extraction phase and need to be addressed, such as repetition of expressions, omission of logical relationships, and deviations in logical propositions and expressions. 
\section*{Acknowledgements}

This work is supported by National Natural Science Foundation of China (No. 62322201 and U23B2020).


\bibliography{bibfile}

\begin{thebibliography}{34}
\providecommand{\natexlab}[1]{#1}

\bibitem[{Achiam et~al.(2023)Achiam, Adler, Agarwal, Ahmad, Akkaya, Aleman, Almeida, Altenschmidt, Altman, Anadkat et~al.}]{openai2024gpt4}
Josh Achiam, Steven Adler, Sandhini Agarwal, Lama Ahmad, Ilge Akkaya, Florencia~Leoni Aleman, Diogo Almeida, Janko Altenschmidt, Sam Altman, Shyamal Anadkat, et~al. 2023.
\newblock Gpt-4 technical report.
\newblock \emph{arXiv preprint arXiv:2303.08774}.

\bibitem[{Anil et~al.(2023)Anil, Dai, Firat, Johnson, Lepikhin, Passos, Shakeri, Taropa, Bailey, Chen et~al.}]{anil2023palm}
Rohan Anil, Andrew~M Dai, Orhan Firat, Melvin Johnson, Dmitry Lepikhin, Alexandre Passos, Siamak Shakeri, Emanuel Taropa, Paige Bailey, Zhifeng Chen, et~al. 2023.
\newblock Palm 2 technical report.
\newblock \emph{arXiv preprint arXiv:2305.10403}.

\bibitem[{Arkoudas(2023)}]{arkoudas2023gpt}
Konstantine Arkoudas. 2023.
\newblock Gpt-4 can't reason.
\newblock \emph{arXiv preprint arXiv:2308.03762}.

\bibitem[{Bao et~al.(2024)Bao, Zhang, Yang, Wang, and Zhang}]{bao2024llms}
Guangsheng Bao, Hongbo Zhang, Linyi Yang, Cunxiang Wang, and Yue Zhang. 2024.
\newblock Llms with chain-of-thought are non-causal reasoners.
\newblock \emph{arXiv preprint arXiv:2402.16048}.

\bibitem[{Besta et~al.(2024)Besta, Blach, Kubicek, Gerstenberger, Podstawski, Gianinazzi, Gajda, Lehmann, Niewiadomski, Nyczyk et~al.}]{besta2024graph}
Maciej Besta, Nils Blach, Ales Kubicek, Robert Gerstenberger, Michal Podstawski, Lukas Gianinazzi, Joanna Gajda, Tomasz Lehmann, Hubert Niewiadomski, Piotr Nyczyk, et~al. 2024.
\newblock Graph of thoughts: Solving elaborate problems with large language models.
\newblock In \emph{Proceedings of the AAAI Conference on Artificial Intelligence}, volume~38, pages 17682--17690.

\bibitem[{B{\"u}ning and Lettmann(1999)}]{buning1999propositional}
Hans~Kleine B{\"u}ning and Theodor Lettmann. 1999.
\newblock \emph{Propositional logic: deduction and algorithms}, volume~48.
\newblock Cambridge University Press.

\bibitem[{Clark et~al.(2021)Clark, Tafjord, and Richardson}]{clark2021transformers}
Peter Clark, Oyvind Tafjord, and Kyle Richardson. 2021.
\newblock Transformers as soft reasoners over language.
\newblock In \emph{Proceedings of the Twenty-Ninth International Conference on International Joint Conferences on Artificial Intelligence}, pages 3882--3890.

\bibitem[{Han et~al.(2022)Han, Schoelkopf, Zhao, Qi, Riddell, Benson, Sun, Zubova, Qiao, Burtell et~al.}]{han2022folio}
Simeng Han, Hailey Schoelkopf, Yilun Zhao, Zhenting Qi, Martin Riddell, Luke Benson, Lucy Sun, Ekaterina Zubova, Yujie Qiao, Matthew Burtell, et~al. 2022.
\newblock Folio: Natural language reasoning with first-order logic.
\newblock \emph{arXiv preprint arXiv:2209.00840}.

\bibitem[{Kazemi et~al.(2022)Kazemi, Kim, Bhatia, Xu, and Ramachandran}]{kazemi2022lambada}
Mehran Kazemi, Najoung Kim, Deepti Bhatia, Xin Xu, and Deepak Ramachandran. 2022.
\newblock Lambada: Backward chaining for automated reasoning in natural language.
\newblock \emph{arXiv preprint arXiv:2212.13894}.

\bibitem[{Kojima et~al.(2022)Kojima, Gu, Reid, Matsuo, and Iwasawa}]{kojima2022large}
Takeshi Kojima, Shixiang~Shane Gu, Machel Reid, Yutaka Matsuo, and Yusuke Iwasawa. 2022.
\newblock Large language models are zero-shot reasoners.
\newblock \emph{Advances in neural information processing systems}, 35:22199--22213.

\bibitem[{Lanham et~al.(2023)Lanham, Chen, Radhakrishnan, Steiner, Denison, Hernandez, Li, Durmus, Hubinger, Kernion et~al.}]{lanham2023measuring}
Tamera Lanham, Anna Chen, Ansh Radhakrishnan, Benoit Steiner, Carson Denison, Danny Hernandez, Dustin Li, Esin Durmus, Evan Hubinger, Jackson Kernion, et~al. 2023.
\newblock Measuring faithfulness in chain-of-thought reasoning.
\newblock \emph{arXiv preprint arXiv:2307.13702}.

\bibitem[{Lightman et~al.(2023)Lightman, Kosaraju, Burda, Edwards, Baker, Lee, Leike, Schulman, Sutskever, and Cobbe}]{lightman2023lets}
Hunter Lightman, Vineet Kosaraju, Yura Burda, Harri Edwards, Bowen Baker, Teddy Lee, Jan Leike, John Schulman, Ilya Sutskever, and Karl Cobbe. 2023.
\newblock Let's verify step by step.
\newblock \emph{arXiv preprint arXiv:2305.20050}.

\bibitem[{Liu et~al.(2023)Liu, Ning, Teng, Liu, Zhou, and Zhang}]{liu2023evaluating}
Hanmeng Liu, Ruoxi Ning, Zhiyang Teng, Jian Liu, Qiji Zhou, and Yue Zhang. 2023.
\newblock Evaluating the logical reasoning ability of chatgpt and gpt-4.
\newblock \emph{arXiv preprint arXiv:2304.03439}.

\bibitem[{Liu et~al.(2020)Liu, Cui, Liu, Huang, Wang, and Zhang}]{liu2020logiqa}
Jian Liu, Leyang Cui, Hanmeng Liu, Dandan Huang, Yile Wang, and Yue Zhang. 2020.
\newblock Logiqa: A challenge dataset for machine reading comprehension with logical reasoning.
\newblock \emph{arXiv preprint arXiv:2007.08124}.

\bibitem[{Lyu et~al.(2023)Lyu, Havaldar, Stein, Zhang, Rao, Wong, Apidianaki, and Callison-Burch}]{lyu2023faithful}
Qing Lyu, Shreya Havaldar, Adam Stein, Li~Zhang, Delip Rao, Eric Wong, Marianna Apidianaki, and Chris Callison-Burch. 2023.
\newblock Faithful chain-of-thought reasoning.
\newblock \emph{arXiv preprint arXiv:2301.13379}.

\bibitem[{Nye et~al.(2021)Nye, Andreassen, Gur-Ari, Michalewski, Austin, Bieber, Dohan, Lewkowycz, Bosma, Luan et~al.}]{nye2021show}
Maxwell Nye, Anders~Johan Andreassen, Guy Gur-Ari, Henryk Michalewski, Jacob Austin, David Bieber, David Dohan, Aitor Lewkowycz, Maarten Bosma, David Luan, et~al. 2021.
\newblock Show your work: Scratchpads for intermediate computation with language models.
\newblock \emph{arXiv preprint arXiv:2112.00114}.

\bibitem[{Olausson et~al.(2023)Olausson, Gu, Lipkin, Zhang, Solar-Lezama, Tenenbaum, and Levy}]{olausson2023linc}
Theo~X Olausson, Alex Gu, Benjamin Lipkin, Cedegao~E Zhang, Armando Solar-Lezama, Joshua~B Tenenbaum, and Roger Levy. 2023.
\newblock Linc: A neurosymbolic approach for logical reasoning by combining language models with first-order logic provers.
\newblock \emph{arXiv preprint arXiv:2310.15164}.

\bibitem[{Pan et~al.(2023)Pan, Albalak, Wang, and Wang}]{pan2023logic}
Liangming Pan, Alon Albalak, Xinyi Wang, and William~Yang Wang. 2023.
\newblock Logic-lm: Empowering large language models with symbolic solvers for faithful logical reasoning.
\newblock In \emph{The 2023 Conference on Empirical Methods in Natural Language Processing}.

\bibitem[{Shum et~al.(2023)Shum, Diao, and Zhang}]{shum2023automatic}
KaShun Shum, Shizhe Diao, and Tong Zhang. 2023.
\newblock Automatic prompt augmentation and selection with chain-of-thought from labeled data.
\newblock \emph{arXiv preprint arXiv:2302.12822}.

\bibitem[{Tafjord et~al.(2021)Tafjord, Dalvi, and Clark}]{tafjord2021proofwriter}
Oyvind Tafjord, Bhavana Dalvi, and Peter Clark. 2021.
\newblock Proofwriter: Generating implications, proofs, and abductive statements over natural language.
\newblock In \emph{Findings of the Association for Computational Linguistics: ACL-IJCNLP 2021}, pages 3621--3634.

\bibitem[{Touvron et~al.(2023)Touvron, Martin, Stone, Albert, Almahairi, Babaei, Bashlykov, Batra, Bhargava, Bhosale et~al.}]{touvron2023llama2}
Hugo Touvron, Louis Martin, Kevin Stone, Peter Albert, Amjad Almahairi, Yasmine Babaei, Nikolay Bashlykov, Soumya Batra, Prajjwal Bhargava, Shruti Bhosale, et~al. 2023.
\newblock Llama 2: Open foundation and fine-tuned chat models.
\newblock \emph{arXiv preprint arXiv:2307.09288}.

\bibitem[{Turpin et~al.(2024)Turpin, Michael, Perez, and Bowman}]{turpin2024language}
Miles Turpin, Julian Michael, Ethan Perez, and Samuel Bowman. 2024.
\newblock Language models don't always say what they think: unfaithful explanations in chain-of-thought prompting.
\newblock \emph{Advances in Neural Information Processing Systems}, 36.

\bibitem[{Wan et~al.(2024)Wan, Wang, Yang, Yuan, Huang, He, Jiao, and Lyu}]{wan2024b}
Yuxuan Wan, Wenxuan Wang, Yiliu Yang, Youliang Yuan, Jen-tse Huang, Pinjia He, Wenxiang Jiao, and Michael~R Lyu. 2024.
\newblock A \& b== b \& a: Triggering logical reasoning failures in large language models.
\newblock \emph{arXiv preprint arXiv:2401.00757}.

\bibitem[{Wang et~al.(2022{\natexlab{a}})Wang, Zhong, Tang, Wei, Fan, Jiang, Zhou, and Duan}]{wang2022logic}
Siyuan Wang, Wanjun Zhong, Duyu Tang, Zhongyu Wei, Zhihao Fan, Daxin Jiang, Ming Zhou, and Nan Duan. 2022{\natexlab{a}}.
\newblock Logic-driven context extension and data augmentation for logical reasoning of text.
\newblock In \emph{Findings of the Association for Computational Linguistics: ACL 2022}, pages 1619--1629.

\bibitem[{Wang et~al.(2022{\natexlab{b}})Wang, Wei, Schuurmans, Le, Chi, Narang, Chowdhery, and Zhou}]{wang2022self}
Xuezhi Wang, Jason Wei, Dale Schuurmans, Quoc~V Le, Ed~H Chi, Sharan Narang, Aakanksha Chowdhery, and Denny Zhou. 2022{\natexlab{b}}.
\newblock Self-consistency improves chain of thought reasoning in language models.
\newblock In \emph{The Eleventh International Conference on Learning Representations}.

\bibitem[{Wei et~al.(2022)Wei, Wang, Schuurmans, Bosma, Xia, Chi, Le, Zhou et~al.}]{wei2023chainofthought}
Jason Wei, Xuezhi Wang, Dale Schuurmans, Maarten Bosma, Fei Xia, Ed~Chi, Quoc~V Le, Denny Zhou, et~al. 2022.
\newblock Chain-of-thought prompting elicits reasoning in large language models.
\newblock \emph{Advances in neural information processing systems}, 35:24824--24837.

\bibitem[{Yao et~al.(2024)Yao, Yu, Zhao, Shafran, Griffiths, Cao, and Narasimhan}]{yao2024tree}
Shunyu Yao, Dian Yu, Jeffrey Zhao, Izhak Shafran, Tom Griffiths, Yuan Cao, and Karthik Narasimhan. 2024.
\newblock Tree of thoughts: Deliberate problem solving with large language models.
\newblock \emph{Advances in Neural Information Processing Systems}, 36.

\bibitem[{Ye et~al.(2024)Ye, Chen, Dillig, and Durrett}]{ye2024satlm}
Xi~Ye, Qiaochu Chen, Isil Dillig, and Greg Durrett. 2024.
\newblock Satlm: Satisfiability-aided language models using declarative prompting.
\newblock \emph{Advances in Neural Information Processing Systems}, 36.

\bibitem[{Yifan et~al.(2024)Yifan, Jingqin, Yang, and Andrew}]{zhang2024cumulative}
Zhang Yifan, Yang Jingqin, Yuan Yang, and Yao Andrew, Chi-Chih. 2024.
\newblock Cumulative reasoning with large language models.
\newblock \emph{arXiv preprint arXiv:2308.04371}.

\bibitem[{Yu et~al.(2020)Yu, Jiang, Dong, and Feng}]{yu2020reclor}
Weihao Yu, Zihang Jiang, Yanfei Dong, and Jiashi Feng. 2020.
\newblock Reclor: A reading comprehension dataset requiring logical reasoning.
\newblock \emph{arXiv preprint arXiv:2002.04326}.

\bibitem[{Zelikman et~al.(2022)Zelikman, Wu, Mu, and Goodman}]{eric2022star}
Eric Zelikman, Yuhuai Wu, Jesse Mu, and Noah Goodman. 2022.
\newblock Star: Bootstrapping reasoning with reasoning.
\newblock \emph{Advances in Neural Information Processing Systems}, 35:15476--15488.

\bibitem[{Zhang et~al.(2024)Zhang, Du, Cao, Fu, and Liu}]{zhang2024guiding}
Yizhou Zhang, Lun Du, Defu Cao, Qiang Fu, and Yan Liu. 2024.
\newblock Guiding large language models with divide-and-conquer program for discerning problem solving.
\newblock \emph{arXiv preprint arXiv:2402.05359}.

\bibitem[{Zhou et~al.(2022{\natexlab{a}})Zhou, Sch{\"a}rli, Hou, Wei, Scales, Wang, Schuurmans, Cui, Bousquet, Le et~al.}]{zhou2022least}
Denny Zhou, Nathanael Sch{\"a}rli, Le~Hou, Jason Wei, Nathan Scales, Xuezhi Wang, Dale Schuurmans, Claire Cui, Olivier Bousquet, Quoc Le, et~al. 2022{\natexlab{a}}.
\newblock Least-to-most prompting enables complex reasoning in large language models.
\newblock \emph{arXiv preprint arXiv:2205.10625}.

\bibitem[{Zhou et~al.(2022{\natexlab{b}})Zhou, Muresanu, Han, Paster, Pitis, Chan, and Ba}]{zhou2022large}
Yongchao Zhou, Andrei~Ioan Muresanu, Ziwen Han, Keiran Paster, Silviu Pitis, Harris Chan, and Jimmy Ba. 2022{\natexlab{b}}.
\newblock Large language models are human-level prompt engineers.
\newblock \emph{arXiv preprint arXiv:2211.01910}.

\end{thebibliography}

\clearpage
\appendix

\section{Comparative Study of States in ToT and LoT+ToT}
\label{appendix:tot-vs-lot+tot}

\begin{figure*}[htb]
    \centering
    \includegraphics[width=1\linewidth]{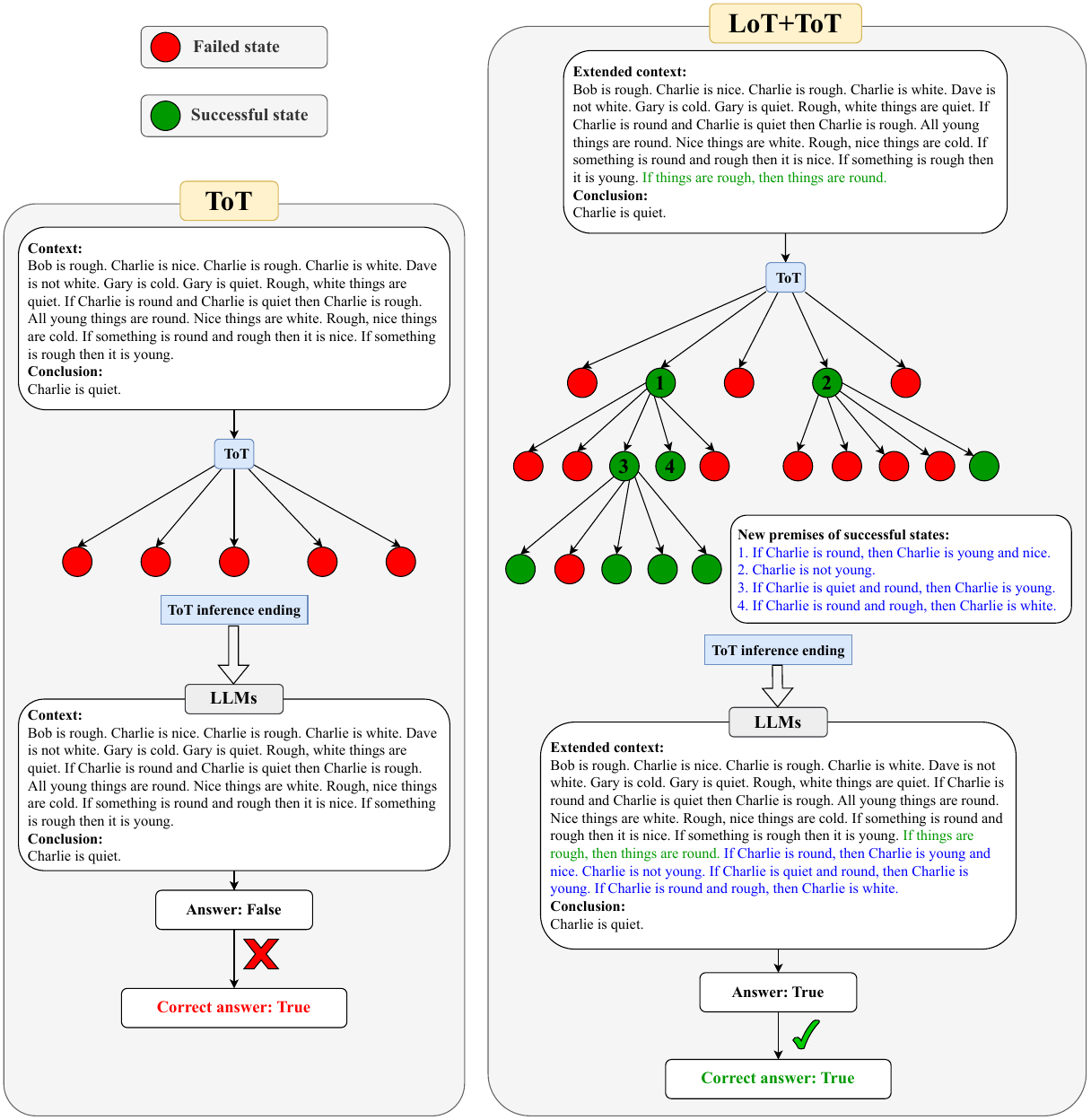}
    \caption{\textbf{Case study of state exploration in ToT and LoT+ToT.}}
    \label{fig:tot-lot-workflow}
\end{figure*}

We present a comprehensive analysis of an illustrative example, comparing the exploration of states when utilizing ToT and LoT+ToT. In Figure \ref{fig:tot-lot-workflow}, we can observe that in LoT+ToT, LoT generates the logical description \textit{"If things are rough, then things are round"}, from which ToT further generates 4 successful states. The corresponding premises are: (1)\textit{"If Charlie is round, then Charlie is young and nice"}, (2)\textit{"Charlie is not young"}, (3)\textit{"If Charlie is quiet and round, then Charlie is young"}, (4)\textit{"If Charlie is round and rough, then Charlie is white"}. Subsequently, the generated information by the LoT and ToT serves as an enhancement to the input prompt, enabling LLMs to produce correct results. Compared to using ToT alone, the logical description generated by LoT enables ToT to generate an additional four successful states, which leads to the correct results. This indicates that LoT enhances the total number of states as well as the number of successful states, thereby expanding the reasoning space and improving the accuracy of ToT reasoning.

\section{An Error Case of LoT}
\label{appendix:limitation-lot}

\begin{figure*}[h]
    \centering
    \includegraphics[width=1\linewidth]{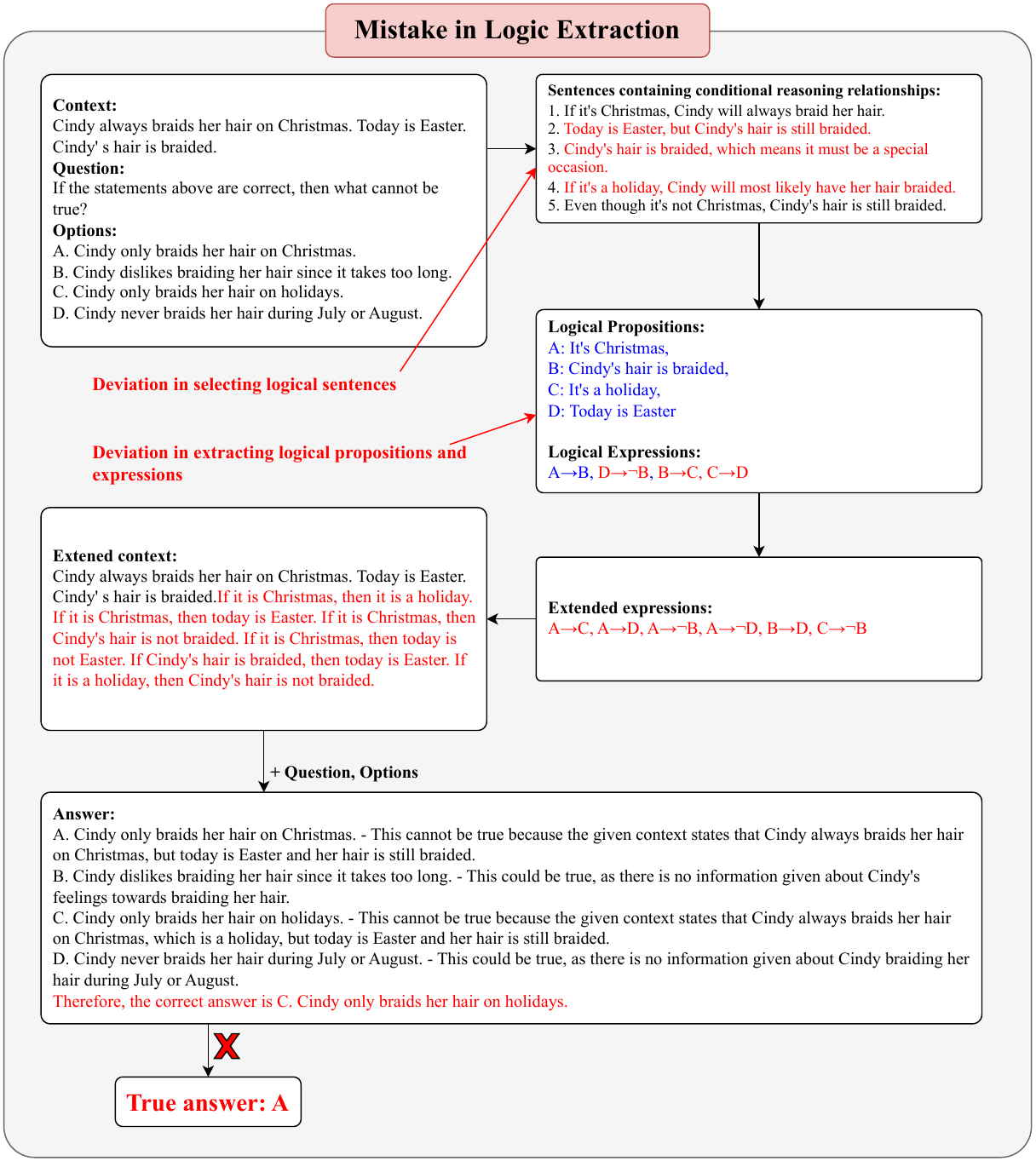}
    \caption{\textbf{Errors in Logic Extraction using LoT.}}
    \label{fig:logic-extraction-failure}
\end{figure*}

Figure \ref{fig:logic-extraction-failure} illustrates an instance of inaccuracies in extracting logical information during the Logic Extraction process, leading to erroneous logical expressions and errors in the final outcome. When LoT selects sentences with logical relationships, there are biases in the information extracted by LLMs. The sentences \textit{"Today is Easter, but Cindy’s hair is still braided"}, \textit{"Cindy’s hair is braided, which means it must be a special occasion"}, and \textit{"If it’s a holiday, Cindy will most likely have her hair braided"} all exhibit inaccuracies. Subsequently, extracted logical expressions, such as $D \to \neg B$, $B \to C$, $C \to D$ exhibit errors. These accumulated errors result in erroneous generated logical descriptions and incorrect final outcomes.

\section{An Example of Overlap Capabilities between CoT and LoT}
\label{appendix:cot-lot}

The following example illustrates the overlap in capabilities between CoT and LoT. In the example, LoT first extracts propositions $A$, $B$, and $C$ from the context and identifies the relationships $A \to B$ and $B \to C$. Then, it extends to a new expression $A \to C$. This new expression is translated into additional logical information \textit{"If a person reads a book, that person becomes smarter"}. This logical information directly links \textit{"Harry read the book"} and \textit{"become smarter"} in the context, helping LLMs correctly infer the answer. CoT's reasoning process involves first deriving proposition $B$ from proposition $A$ based on the second sentence, then deriving proposition $C$ from proposition $B$ based on the third sentence, ultimately arriving at the answer. We can see that both CoT and LoT handle this problem by linking conditional statements and reasoning step by step, indicating that CoT and LoT sometimes have overlapping capabilities.
\subsection{An Example of Overlap Capabilities Between CoT and LoT}
\lstset{style=mystyle}

\begin{lstlisting}
<@\textcolor{brown}{\# Context:}@>
1. Books contain tons of knowledge.
2. When a person reads a book, that person gains knowledge.
3. If a person gains knowledge, they become smarter.
4. Harry read the book "Walden" by Henry Thoreau.

<@\textcolor{brown}{\# Whether this inference is correct:}@>
Harry is smarter than before.
-------------------------------------
<@\textcolor{brown}{\# LoT:}@>
<@\textcolor{brown}{\#\# Logic Extraction:}@>
2. When a person reads a book, that person gains knowledge.
3. If a person gains knowledge, they become smarter.
----
<@$A$@>: a person reads a book, <@$B$@>: person gains knowledge, <@$C$@>: become smarter
<@$A \to B$, $B \to C$@>

<@\textcolor{brown}{\#\# Logic Extension:}@>
<@$A \to C$@>

<@\textcolor{brown}{\#\# Logic Translation:}@>
If a person reads a book, that person become smarter.

<@\textcolor{brown}{\#\# Extended context:}@>
1. Books contain tons of knowledge.
2. When a person reads a book, that person gains knowledge.
3. If a person gains knowledge, they become smarter.
4. Harry read the book "Walden" by Henry Thoreau.
5. If a person reads a book, that person become smarter.

<@\textcolor{brown}{\#\# LLM Answer:}@>
<@\textcolor{deepgreen}{\emph{Correct}}@>
-------------------------------------
<@\textcolor{brown}{\# CoT:}@>
Let's think step by step:
Given that Harry read the book "Walden" by Henry Thoreau, it can be concluded that he gained knowledge from reading the book. 
Therefore, based on the context provided, it is reasonable to conclude that Harry is smarter than before.

<@\textcolor{brown}{\#\# LLM Answer:}@>
<@\textcolor{deepgreen}{\emph{Correct}}@>
\end{lstlisting}


\begin{table*}[t]
    \centering
    \resizebox{\linewidth}{!}{
    \begin{tabular}{@{}lccc@{}}
        \toprule
        Direct $\rightarrow$ LoT         & Right Extraction(\%) & Wrong Extraction(\%) & Sum(\%)    \\ \midrule
        Answer Right $\rightarrow$ Wrong & 5.19             & 6.67             & 11.85  \\
        Answer Wrong $\rightarrow$ Right & 8.15             & 10.37            & 18.52  \\
        Answer Right $\rightarrow$ Right & 38.51            & 19.26            & 57.78  \\
        Answer Wrong $\rightarrow$ Wrong & 8.15             & 3.70             & 11.85  \\ \midrule
        SUM                   & 60.00               & 40.00               & 100.00 \\ 
        \bottomrule
    \end{tabular}
    }
    \caption{\textbf{Direct → LoT: The impact of Logical Extraction phase.}}
    \label{The impact of logical extraction}
\end{table*}

\begin{table*}[t]
    \centering
    \resizebox{\linewidth}{!}{
    \begin{tabular}{@{}lccc@{}}
        \toprule
        Direct $\rightarrow$ LoT         & Right Translation(\%) & Wrong Translation (\%) & Sum(\%)    \\ \midrule
        Answer Right $\rightarrow$ Wrong & 11.11             & 0.74              & 11.85  \\
        Answer Wrong $\rightarrow$ Right & 16.30             & 2.22              & 18.52  \\
        Answer Right $\rightarrow$ Right & 56.30             & 1.48              & 57.78  \\
        Answer Wrong $\rightarrow$ Wrong & 11.11             & 0.74              & 11.85  \\ \midrule
        Sum                   & 94.82             & 5.18              & 100.00 \\ 
        \bottomrule
    \end{tabular}
    }
    \caption{\textbf{Direct $\rightarrow$ LoT: The impact of Logical Translation phase.}}
\end{table*}

\begin{table*}[t]
    \centering
    \resizebox{0.8\linewidth}{!}{
    \begin{tabular}{@{}lccc@{}}
        \toprule
        Direct $\rightarrow$ LoT         & LoT Right(\%) & LoT Wrong(\%) & Sum(\%)    \\ \midrule
        Answer Right $\rightarrow$ Wrong & 5.19      & 6.67      & 11.85  \\
        Answer Wrong $\rightarrow$ Right & 6.67      & 11.85     & 18.52  \\
        Answer Right $\rightarrow$ Right & 37.78     & 20.00     & 57.78  \\
        Answer Wrong $\rightarrow$ Wrong & 8.15      & 3.70      & 11.85  \\ \midrule
        Sum                   & 57.78     & 42.22     & 100.00 \\ 
        \bottomrule
    \end{tabular}
    }
    \caption{\textbf{Direct → LoT: The overall impact of LoT.}}
\end{table*}

\section{The in-depth analysis of LoT}
LoTh comprises three phases: Logical Extraction, Logical Expansion, and Logical Translation. In this study, we statistically compare the experimental outcomes of Direct and LoT utilizing the \texttt{GPT-3.5-turbo-0125} on the FOLIO dataset. Our analysis is presented in Table, each corresponding to the correctness of the Logical Extraction phase, the Logical Translation phase, and the overall impact of the LoT on changes in the answers.
Within the entire sample set, it was observed that 40\% of the samples encountered errors during the logical extraction phase. No errors were recorded during logical expansion. Additionally, 5.18\% of the samples experienced errors in the logical translation stage. Overally, 42.22\% of the samples exhibited at least one error across the three steps.

The data displayed in the table indicates the proportion of samples with changes in results due to specific errors relative to the entire sample set. For example, the data in the first row and first column of Table \ref{The impact of logical extraction} indicate that when the logical extraction process is executed correctly, 5.19\% of the total sample can obtain the correct answer using the direct method, while using the LoT method yields the incorrect answer.

\section{Full Set of Prompts}
\label{appendix:full-set-of-prompts}

\subsection{Logic Extraction Prompt in LoT}
\label{appendix:logic-extraction-prompt}

\begin{lstlisting}
<@\textcolor{brown}{\# Prompt for ReClor and LogiQA:}@>
Please use uppercase English letters such as A, B, C, etc. to identify all possible propositions. Do not include negative tones such as "not" in the propositions. For example, if the sentence is "It is not bored," you should use "A: bored" to represent it.

Next, for each proposition, use the symbol to represent its negative form. For example, the negative form of proposition A can be expressed as A.

Now, please carefully analyze the context and find causal relationship between propositions seriously. A causal expression is only established when the context directly supports this relationship. Use arrows (<@$\to$@>) to indicate causal relationships, for example, "If A, then B", "B if A" and "A causes B" etc. can be represented as A<@$\to$@>B.

Finally, output propositions and causal expressions.
\end{lstlisting}

\begin{lstlisting}
<@\textcolor{brown}{\# Prompt for RuleTaker, ProofWriter and FOLIO:}@>
Please use uppercase English letters such as A, B, C, etc. to identify all possible propositions. Do not include negative tones such as "not" in the propositions. For example, if the sentence is "It is not bored," you should use "A: bored" to represent it.

Next, for each proposition, use the symbol to represent its negative form. For example, the negative form of proposition A can be expressed as <@$\neg$@>A.

Now, please carefully analyze the context and find causal relationship between propositions. A causal expression is only established when the context directly supports this relationship. Use arrows (<@$\to$@>) to indicate causal relationships, for example, "If A, then B", "B if A" and "A causes B" etc. can be represented as A<@$\to$@>B.

Finally, output propositions and causal expressions.
\end{lstlisting}

\subsection{Logic Translation Prompt in LoT}
\label{appendix:logic-translation-prompt}
\begin{lstlisting}
<@\textcolor{brown}{\# Logical Translation Prompt for All Datasets:}@>
Please use the provided propositions to translate each expression into a complete sentence.

<@$\neg$@>A represents the negation of proposition A, the arrow (<@$\to$@>) represents the causal relationship, and A<@$\to$@>B represents if A, then B.

Only output the sentences in a paragraph! 
\end{lstlisting}

\subsection{SatLM Prompt}
\label{appendix:satlm-prompt}

\begin{lstlisting}
<@\textcolor{brown}{\# SatLM Prompt for LSAT:}@>
Nine different treatments are available for a certain illness: three antibiotics (F, G, and H) three dietary regimens (M, N, and O) and three physical therapies (U, V, and W). For each case of the illness, a doctor will prescribe exactly five of the treatments, in accordance with the following conditions: If two of the antibiotics are prescribed, the remaining antibiotic cannot be prescribed. There must be exactly one dietary regimen prescribed. If O is not prescribed, F cannot be prescribed. If W is prescribed, F cannot be prescribed. G cannot be prescribed if both N and U are prescribed. V cannot be prescribed unless both H and M are prescribed. 

Question: If O is prescribed for a given case, which one of the following is a pair of treatments both of which must also be prescribed for that case?
(A) F, M (B) G, V (C) N, U (D) U, V (E) U, W 

treatments = [F, G, H, M, N, O, U, V, W] 
antibiotics = [F, G, H] 
dietary_regimens = [M, N, O] 
physical_therapies = [U, V, W] 
prescribed = Function(treatments, bool) 

Count([t:treatments], prescribed(t)) == 5 
Count([a:antibiotics], prescribed(a)) <= 2 
Count([d:dietary_regimens], prescribed(d)) == 1 
Implies(Not(prescribed(O)), Not(prescribed(F))) 
Implies(prescribed(W), Not(prescribed(F))) 
Implies(And(prescribed(N), prescribed(U)), Not(prescribed(G))) 
Implies(prescribed(V), And(prescribed(H), prescribed(M))) 

solve(Implies(prescribed(O), And(prescribed(U), prescribed(V)))) # (A) 
solve(Implies(prescribed(O), And(prescribed(G), prescribed(V)))) # (B) 
solve(Implies(prescribed(O), And(prescribed(N), prescribed(U)))) # (C) 
solve(Implies(prescribed(O), And(prescribed(U), prescribed(V)))) # (D) 
solve(Implies(prescribed(O), And(prescribed(U), prescribed(W)))) # (E)
\end{lstlisting}

\subsection{ToT Prompt}
\label{appendix:tot-prompt}

\begin{lstlisting}
<@\textcolor{brown}{\# ToT Prompt for Final Conclusion:}@>
{{#system}}
Suppose you are one of the greatest AI scientists, logicians and mathematicians. Let us think step by step. 
Read and analyze the "Premises" first, then judge whether the "Hypothesis" is True, False.
Please make sure your reasoning is directly deduced from the "Premises" and "Propositions" other than introducing unsourced common knowledge and unsourced information by common sense reasoning.
----
{{/system}}

{{~#each examples}}
{{#user}}
---
"Premises": "{{this.premises}}"
"Hypothesis": "{{this.conclusion}}"
{{/user}}

{{#assistant}}
"Thoughts": "Let us think step by step. From the premises, we can deduce propositions: {{this.propositions}}"
{{/assistant}}
{{#assistant}}
"Reasoning": "Let us think step by step, {{this.reasoning}}"
{{/assistant}}
{{#assistant}}
"Recall the Hypothesis": "{{this.conclusion}}"
{{/assistant}}
{{#assistant}}
"Judgement": "Now we know that the Hypothesis is {{this.judgement}}{{/assistant}}
{{~/each}}

{{#user}}
---
"Premises": "{{premises}}"
"Hypothesis": "{{conclusion}}"
{{/user}}

{{#assistant}}
"Thoughts": "Let us think step by step. From the premises, we can deduce propositions: {{propositions}}"
{{/assistant}}
{{#assistant}}
"Recall the Hypothesis": "{{conclusion}}"
{{/assistant}}
{{#assistant}}
"Reasoning": "Let us think step by step,
{{/assistant}}
{{#assistant}}
{{gen "reasoning" temperature=0.7 max_tokens=300 stop=['textbackslash n']}}{{/assistant}}
{{#assistant}}
"Recall the Hypothesis": "{{conclusion}}"
{{/assistant}}
{{#assistant}}
"Judgement": "Now we know that the Hypothesis is 
{{/assistant}}
{{#assistant}}
{{gen "judgement" temperature=temperature max_tokens=1 stop='textbackslash \n'}}
{{/assistant}}
\end{lstlisting}

\begin{lstlisting}
<@\textcolor{brown}{\# ToT Prompt for Generate Proposition:}@>            
{{#system}}
Suppose you are one of the greatest AI scientists, logicians and mathematicians. Let us think step by step.  Please use Logical Reasoning Rules(LRR) to deduce a "Proposition" from two given "Premises" and the proposition does not include "if". Logical Reasoning Rules(LRR):  1. "Two premises": "If A,then B. A is true." then "Proposition": "B is true." 2. "Two premises": "If A,then B. B is not true." then "Proposition": "A is not true" 3. "Two premises": "A is either C or D. A is not C." then "Proposition": "A is D."    Please make sure that the "Proposition" is logically correct.    Please make sure that the "Proposition" is not a duplicate of the "Premises".    Please make sure your reasoning is directly deduced from the "Premises" and "Propositions" other than introducing unsourced common knowledge and unsourced information by common sense reasoning.    Please remember that your "Proposition" should be useful to determine whether the "Hypothesis" is True, False.    
----{{#system}}  

{{~#each examples}} 
{{#user}} ---  ``Premises": "{{this.premises}}" We want to deduce more propositions to determine the correctness of the following "Hypothesis": ``Hypothesis": "{{this.conclusion}}" Can you deduce a new "Proposition" from at least two given "Premises"? 
{{#user}}  
{{#assistant}}
"Proposition": "{{this.propositions}}"
{{/assistant}} 
{{~/each}}  
{{#user}} ---  Premises": "{{this.premises}}" We want to deduce more propositions to determine the correctness of the following "Hypothesis": ``Hypothesis": "{{this.conclusion}}" Can you deduce a new "Proposition" from at least two given "Premises"? 
{{#user}}  
{{#assistant}}
"Proposition": "
{{/assistant}} 
{{#assistant}}
{{gen "proposition" temperature=temperature max_tokens=50 stop='textbackslash \n'}}
{{/assistant}} 
\end{lstlisting}

\begin{lstlisting}
<@\textcolor{brown}{\# ToT Prompt for Validate Deduction:}@>        
{{#system}}
Suppose you are one of the greatest AI scientists, logicians and mathematicians. Let us think step by step. 
Please use the Logical Reasoning Rules(LRR) to determine whether the deduction of the given "Premises" to a "Proposition" is valid or not, reply with True or False.
Logical Reasoning Rules(LRR):
1. "Two premises": "If A,then B. A is true." then "Proposition": "B is true."
2. "Two premises": "If A,then B. If B,then C." then "Proposition": "If A, then C."
3. "Two premises": "If A,then B. B is not true." then "Proposition": "A is not true"
4. "Two premises": "A is either C or D. A is not C." then "Proposition": "A is D."
----{{/system}}
{{~#each examples}}
{{#user}}
---
"Premises": "{{this.premises}}"
"Proposition": "{{this.propositions}}"
{{/user}}

{{#assistant}}
"Judgement": "Is this deduction valid? {{this.validation}}"
{{/assistant}}
{{~/each}}

{{#user}}
---
"Premises": "{{premises}}"
"Proposition": "{{propositions}}"
{{/user}}

{{#assistant}}
"Judgement": "Is this deduction valid? 
{{/assistant}}
{{#assistant}}
{{gen "validation" temperature=temperature max_tokens=1 stop='textbackslash \n'}}
{{/assistant}}
\end{lstlisting}

\begin{lstlisting}
<@\textcolor{brown}{\# ToT Prompt for Sourced Deduction:}@> 
{{#system}}Suppose you are one of the greatest AI scientists, logicians and mathematicians. Let us think step by step. 
Please determine whether the "Proposition" is directly deduced from the "Premises" with certainty other than introducing unsourced information by common sense reasoning, reply with True or False.
----
{{/system}}

{{~#each examples}}
{{#user}}
---
"Premises": "{{this.premises}}"
"Proposition": "{{this.propositions}}"
{{/user}}

{{#assistant}}
"Judgement": "Is this proposition directly deduced from the premises? {{this.sourced}}"
{{/assistant}}
{{~/each}}

{{#user}}
---
"Premises": "{{premises}}"
"Proposition": "{{propositions}}"
{{/user}}

{{#assistant}}
"Judgement": "Is this proposition directly deduced from the premises? {{/assistant}}
{{#assistant}}
{{gen "sourced" temperature=temperature max_tokens=1 stop='textbackslash \n'}}{{/assistant}}
\end{lstlisting}

\end{document}